\documentclass[runningheads]{llncs}

\usepackage{eccv}

\usepackage{eccvabbrv}

\usepackage{graphicx}
\usepackage{booktabs}
\usepackage{multirow}
\usepackage{amsmath}
\usepackage{amssymb}
\usepackage{xcolor}

\usepackage[accsupp]{axessibility}  

\usepackage{hyperref}

\begin{document}

\title{BoxTuning: Directly Injecting the Object Box\texorpdfstring{\\}{ }for Multimodal Model Fine-Tuning}

\titlerunning{BoxTuning}

\author{
Zekun Qian\inst{1}\hspace{1em}
Ruize Han\inst{2}\textsuperscript{\textdagger}\hspace{1em}
Wei Feng\inst{1}
}
\authorrunning{Z. Qian et al.}
\institute{
School of Computer Science and Technology, Tianjin University, Tianjin, China\\
\email{\{clarkqian,wfeng\}@tju.edu.cn}
\and
Faculty of Computer Science and Artificial Intelligence, Shenzhen University of Advanced Technology, Shenzhen, China\\
\email{hanruize@suat-sz.edu.cn}
}

\maketitle
\renewcommand{\thefootnote}{\fnsymbol{footnote}}
\footnotetext[4]{Corresponding author.}

\begin{abstract}
Object-level spatial-temporal understanding is essential for video question answering, yet existing multimodal large language models (MLLMs) encode frames holistically and lack explicit mechanisms for fine-grained object grounding.
Recent work addresses this by serializing bounding box coordinates as text tokens, but this text-coordinate paradigm suffers from a fundamental \textit{modality mismatch}: object information is inherently visual, yet encoding it as text incurs a high token cost that forces aggressive temporal downsampling.
We propose \textit{BoxTuning}, which resolves this mismatch by injecting object spatial-temporal information directly into the visual modality.
Colored bounding boxes and trajectory trails are rendered onto video frames as visual prompts, with only a concise color-to-object legend retained as text.
This reduces the token cost significantly, achieving 87--93\% text token reduction in practice. It also preserves full temporal resolution, where the trajectory trails further encode inter-frame motion direction and speed within each keyframe, recovering fine-grained dynamics that text-coordinate methods are forced to discard.
Experimental results on five video QA benchmarks (CLEVRER, Perception Test, STAR, NExT-QA, IntentQA) show that BoxTuning surpasses text-coordinate baselines on spatially oriented tasks and nearly eliminates the accuracy degradation observed on reasoning-centric tasks, establishing visual prompting as a more natural and efficient paradigm for conveying object information to video MLLMs.
\end{abstract}

\section{Introduction}
\label{sec:intro}

Video understanding requires models to jointly reason over appearance, motion, and interaction across time, a challenge that becomes especially acute when the scene contains multiple interacting agents and objects~\cite{yi2020clevrer,wu2021star,xiao2021nextqa,li2023intentqa}.
A broad range of video question answering tasks demand \textit{object-level} grounding: identifying \textit{what} objects are present, \textit{where} they are located, and \textit{how} they move and interact over time.
Such understanding goes beyond simple video recognition (classification). For example, answering ``\textit{Will the ball hit the red cylinder?}'' requires tracking spatial trajectories across frames~\cite{yi2020clevrer}; answering ``\textit{Why did the man pick up the bag?}'' requires grounding the agent and object through a causal chain~\cite{li2023intentqa};
answering ``\textit{What does the person do after sitting down?}'' requires localizing the actor and linking actions to objects in sequence~\cite{wu2021star}.

For visual understanding, multimodal large language models (MLLMs)~\cite{liu2023llava,li2023blip2,dai2023instructblip,alayrac2022flamingo} have achieved remarkable progress by bridging vision encoders with large language models.
Recent efforts have extended these architectures to video, producing a family of video MLLMs~\cite{cheng2024videollama2,zhang2023videollama,maaz2024videochatgpt,li2024mvbench,jin2024chatunivi} that process temporal video sequences.
Despite the progress, current MLLMs struggle to handle the above object-level understanding tasks.
This is because these models encode each frame \textit{holistically} through a vision encoder (\eg, CLIP ViT~\cite{radford2021clip}), producing frame-level feature representations that lack explicit information about individual objects.
As a result, the language model must implicitly infer object identities, locations, and movements from distributed visual features, a task that becomes increasingly difficult as the number of objects and the complexity of their interactions grow.
In general, a common bottleneck of MLLMs is the lack of an explicit, efficient mechanism to communicate the object-level information to the language decoder.

To address this gap, Tang \etal~\cite{tang2025objectmllm} propose ObjectMLLM, which augments the MLLM input with structured bounding box information from an off-the-shelf detector and tracker.
Each object's bounding box coordinates across frames are serialized as text tokens and appended to the prompt (\eg, ``\texttt{Boy: frame\,1 [60 58 70 98] frame\,2 [57 48 70 95]~\ldots}'').
Such a text-coordinate representation proves highly effective on benchmarks that emphasize spatial-temporal object understanding: ObjectMLLM significantly improves accuracy over the baseline VideoLLaMA2~\cite{cheng2024videollama2} on 
synthetic collision reasoning (CLEVRER~\cite{yi2020clevrer}) and real-world perceptual understanding (Perception Test~\cite{patraucean2023perception}, STAR~\cite{wu2021star}).
The success of ObjectMLLM establishes a clear message: \textit{explicit object-level spatial-temporal information is valuable for video understanding}.

\begin{figure}[tb]
	\centering
	\includegraphics[width=\linewidth]{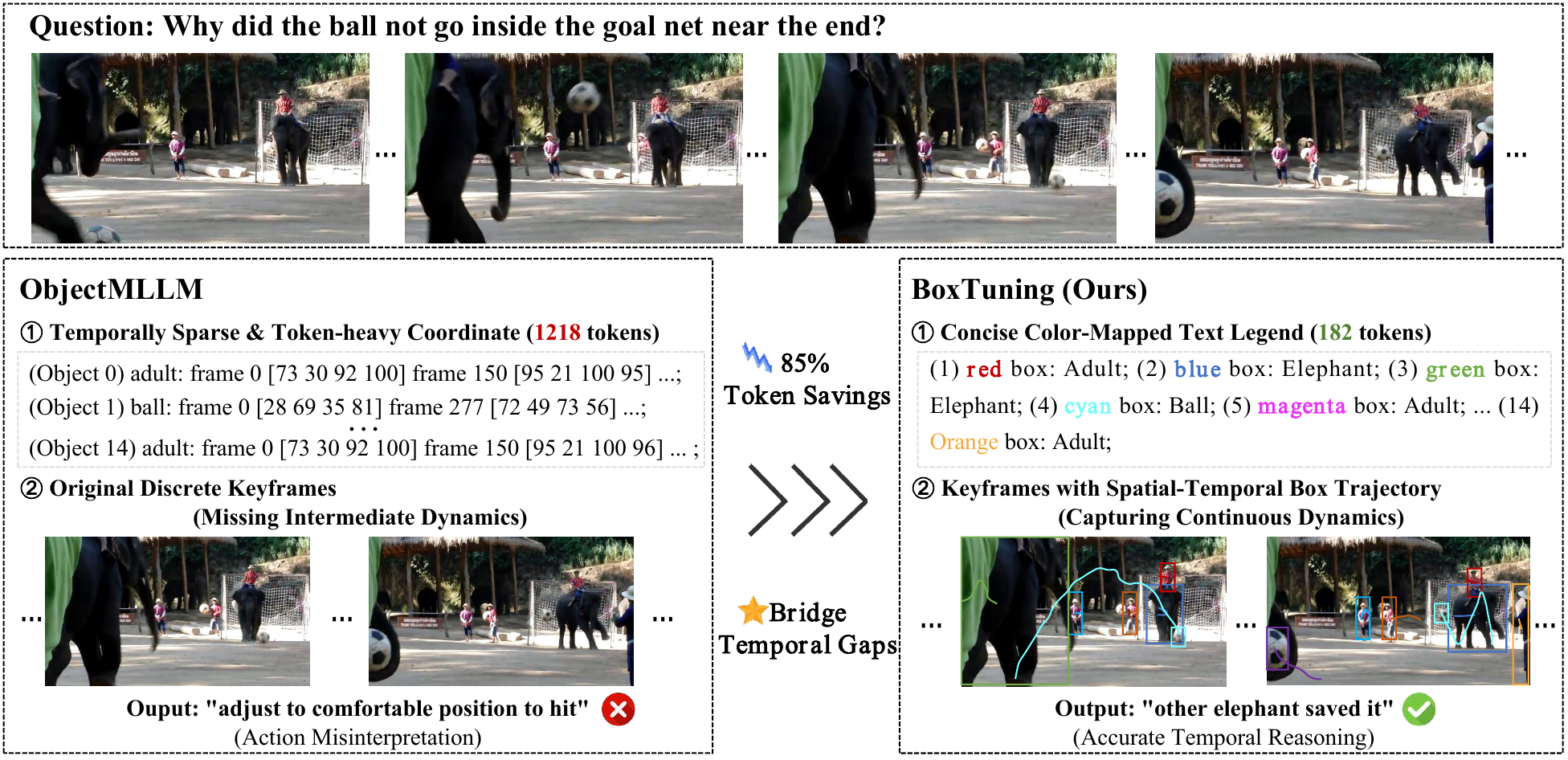}  \vspace{-15pt}
	\caption{Comparison of object-centric representation strategies for video MLLMs. \textit{Left}: ObjectMLLM~\cite{tang2025objectmllm} serializes per-frame bounding box coordinates as text tokens (\eg, ``\texttt{(Object\,0) adult: frame\,0 [73 30 92 100] \ldots}''), producing 1{,}218 tokens for a single video. The original keyframes lack intermediate dynamics, leading to action misinterpretation. \textit{Right}: BoxTuning renders colored bounding boxes and trajectory trails directly on the video frames, with only a concise color-mapped text legend (\eg, ``\texttt{(1) Red box: Adult; (2) Blue box: Elephant; \ldots}'', 182 tokens). The trajectory-augmented keyframes capture continuous dynamics, enabling accurate temporal reasoning. Overall, BoxTuning achieves 85\% token savings (in this case) while bridging temporal gaps that text-coordinate methods miss.}
	\label{fig:teaser}  \vspace{-20pt}
\end{figure}

However, the pure text-coordinate representation introduces a fundamental efficiency bottleneck.
As shown in the left panel of \cref{fig:teaser}, the token cost scales as $\mathcal{O}(N \times T \times C)$, where $N$ is the number of tracked objects, $T$ is the number of sampled frames, and $C \approx 16$--$22$ is the token cost per coordinate entry (\eg, ``\texttt{frame\,1 [60 58 70 98]}'').
To keep the total token count manageable, ObjectMLLM must temporally downsample the bounding box annotations at rates far below the original video frame rate.
For instance, on STAR~\cite{wu2021star} with dense interacting objects, the bounding boxes are sampled at only one frame per 120 original frames (1/120); on NExT-QA~\cite{xiao2021nextqa} the rate is approximately 1/60, and even on the synthetic CLEVRER~\cite{yi2020clevrer} with relatively few objects the rate reaches 1/25.
Such coarse temporal sampling inevitably discards fine-grained motion information that is precisely what the bounding box coordinates are meant to capture.
Even after this aggressive downsampling, the text overhead remains prohibitively large.
The mean box token count still reaches 1{,}184 per sample on NExT-QA, roughly 27$\times$ larger than the question tokens (mean 43 tokens). 
Limited by the computing memory budget, the overlong box representation tokens may overflow the context window, and even adversely affect the visual tokens and other text tokens from the question and answer.


We identify the root cause of these problems as a \textit{modality mismatch}: object spatial information, including locations, sizes, and motion trajectories, is inherently visual in nature, yet the text-coordinate paradigm forces it through the text modality, where each coordinate consumes precious tokens from the limited context budget.
This creates an irreconcilable tension: to provide comprehensive object information, the model needs more text tokens, which makes the context window over saturated.
Meanwhile, the vision encoder processes every sampled frame in its entirety, yet none of the object spatial knowledge is channeled through this visual pathway.
Such being the case, \textit{why not directly encode the object information into the video (visual modality)?}

Based on this insight, we propose a fundamentally new form of object-centric representation for video MLLMs.
Rather than encoding object states as text-based symbolic sequences as in ObjectMLLM~\cite{tang2025objectmllm}, we change the representation medium from the text modality to the visual modality itself, and propose \textit{BoxTuning}. 
It addresses both problems identified above simultaneously, \ie, the excessive text token consumption and the loss of temporal motion information due to downsampling.
As illustrated in the right panel of \cref{fig:teaser}, BoxTuning introduces two complementary visual prompts.
First, we render \textbf{colored bounding boxes} directly onto each video frame, where each tracked object is assigned a distinct color from a predefined palette.
The color serves as the bridge between the visual and textual modalities: the only text required is a concise legend mapping colors to object classes (\eg, ``\texttt{(1) Red box: Adult; (2) Blue box: Elephant; \ldots}''), reducing the text cost from $\mathcal{O}(N \times T \times C)$ to $\mathcal{O}(N \times C')$, where $C'{\approx}9$ tokens per object is far smaller than the per-frame coordinate cost $C{\approx}18$, and more importantly the temporal factor $T$ is entirely eliminated.
Second, to compensate for the temporal information lost between sampled frames, we draw \textbf{temporal trajectory trails} that connect the bounding box centers of each object over recent frames.
This is inspired by Johansson's classical biological motion experiments~\cite{johansson1973biological}, which demonstrated that humans can recognize complex actions from point-light trajectory displays alone, suggesting that trajectory patterns are a powerful and natural encoding of motion.
Our trails explicitly encode motion direction and speed as visual patterns within a single frame, recovering the fine-grained motion cues that the text-coordinate approach is forced to discard through temporal downsampling.

This encoding approach yields three concrete advantages over text-coordinate representations: 1) \textbf{Dramatic token reduction.} The text cost drops from $\mathcal{O}(N \times T \times C)$ to $\mathcal{O}(N \times C')$. In practice, this amounts to an 87\%-93\% reduction, freeing context window capacity for the visual and question/answer text tokens. 
2) \textbf{Lossless temporal resolution.} We can render each object's motion trajectory across all intermediate frames directly onto the sampled keyframes. 
This preserves the fine-grained motion cues that text-coordinate approaches are forced to discard through aggressive temporal downsampling. 
3) \textbf{Perceptual naturalness.} Colored bounding boxes and trajectory trails mirror how humans annotate video in practice. This representation is intuitive for the vision encoder, which is pretrained on natural images containing similar visual patterns, and for human inspection during debugging.

We apply BoxTuning to VideoLLaMA2-7B~\cite{cheng2024videollama2} with LoRA~\cite{hu2022lora}, using the same object detection and tracking pipeline as ObjectMLLM~\cite{tang2025objectmllm}, ensuring that any performance difference stems solely from the representation rather than the object perception quality.
On five benchmarks, CLEVRER~\cite{yi2020clevrer}, Perception Test~\cite{patraucean2023perception}, STAR~\cite{wu2021star}, NExT-QA~\cite{xiao2021nextqa}, and IntentQA~\cite{li2023intentqa}, BoxTuning exceeds ObjectMLLM's accuracy while consuming only about 10\% of the text tokens required by the text-coordinate representation.
Our main contributions are: \vspace{-0pt}
\begin{itemize}
    \item We propose BoxTuning, a brand new visual prompting approach that directly renders spatial-temporal object information on video frames, replacing verbose text-coordinate representations with a visually grounded alternative. To our knowledge, this is the first work to apply anthropomorphic object-level visual prompts to video MLLMs.
    
    \item In BoxTuning, we encode the spatial object bounding boxes and temporal trajectory trails as visual tokens without additional token overhead, and a small number of text legend entries as text tokens. Through systematic analysis of token efficiency, compared to the previous ObjectMLLM, BoxTuning achieves the 87\%--93\% text token reduction.
    
    \item Experimental results on five video QA benchmarks demonstrate significant accuracy improvements based on the baseline MLLMs and consistent improvements compared to ObjectMLLM. Comprehensive ablation studies validate the contribution of each component.
\end{itemize}

\section{Related Work}
\label{sec:related}

\subsection{Video Understanding with MLLMs}

Large language models have demonstrated remarkable capabilities in text understanding and generation, inspiring a growing body of work that extends these models to the video domain~\cite{li2023videochat,maaz2024videochatgpt,zhang2023videollama,cheng2024videollama2,jin2024chatunivi,li2024mvbench}.
A central challenge for video MLLMs, compared to their image counterparts~\cite{liu2023llava,li2023blip2,dai2023instructblip,alayrac2022flamingo}, is how to effectively and efficiently represent the rich spatiotemporal information in videos.
Most video MLLMs adopt a frame-level encoding strategy: pretrained image encoders~\cite{radford2021clip} extract features from sampled frames individually, and these features are concatenated or pooled to form the video representation.
Video-ChatGPT~\cite{maaz2024videochatgpt} applies spatial and temporal pooling over frame features, while Video-LLaMA~\cite{zhang2023videollama} introduces a Video Q-Former for audio-visual alignment.
VideoLLaMA2~\cite{cheng2024videollama2} further improves temporal modeling with a spatial-temporal convolution (STC) connector that captures cross-frame dependencies before projecting to the language space.
Video-LLaVA~\cite{lin2024videollava} unifies image and video representations into a shared feature space through alignment before projection, enabling joint training across modalities.
MVBench~\cite{li2024mvbench} proposes a comprehensive temporal understanding benchmark covering 20 tasks and introduces VideoChat2 with progressive multi-modal training.
Chat-UniVi~\cite{jin2024chatunivi} handles both images and videos with dynamic visual tokens and spatial merging for efficiency, while LLaMA-VID~\cite{li2024llamavid} compresses each frame into two tokens for long-video processing.
LLaVA-OneVision\cite{li2025llavaonevision} demonstrates strong cross-scenario transfer from images to videos within a single model.
An orthogonal line of work sidesteps visual encoders entirely and represents videos as text, using captions~\cite{zeng2023socratic,wang2024vamos} or structured descriptions that can be directly consumed by LLMs.
While such Socratic approaches~\cite{zeng2023socratic} offer interpretability and data efficiency, they tend to miss fine-grained spatiotemporal information such as object trajectories~\cite{tang2025objectmllm}.
Recent studies~\cite{tong2024eyeswideshut,tong2024cambrian} also reveal systematic visual shortcomings of current MLLMs, suggesting that visual representation learning remains an open challenge.
Despite these advances, most video MLLMs encode frames holistically and lack explicit mechanisms for object-level spatial reasoning, limiting their performance on tasks that require understanding individual object interactions and movements.

\subsection{Object-Level Spatial Representations in MLLMs}

Integrating explicit object information into vision-language models has been studied extensively.
Early pretraining methods such as Oscar~\cite{li2020oscar} use object tags detected in images as anchor points to ease cross-modal alignment, and VinVL~\cite{zhang2021vinvl} demonstrates that higher-quality region features from stronger object detectors lead to substantial gains across vision-language tasks.
CoVLM~\cite{li2024covlm} advances this direction by composing visual entities and relationships through communication tokens that enable dynamic interaction between the detection and language systems.
In the era of MLLMs, Shikra~\cite{chen2023shikra} and Kosmos-2~\cite{peng2024kosmos2} encode bounding box coordinates as text tokens for referential dialogue, while GPT4RoI~\cite{zhang2023gpt4roi} extracts region-of-interest features as additional visual inputs.
Osprey~\cite{yuan2024osprey} and GLaMM~\cite{rasheed2024glamm} further extend spatial grounding to pixel-level mask-text alignment.
SpatialVLM~\cite{chen2024spatialvlm} addresses quantitative spatial reasoning by generating Internet-scale 3D spatial QA data.
Karimi Mamaghan \etal~\cite{karimi2025objcentric} conduct an extensive empirical study comparing object-centric representations with foundation model features, finding that applying object-centric inductive biases to foundation models can reduce computational cost while maintaining performance.
These text-based spatial encoding methods extend naturally to video: ObjectMLLM~\cite{tang2025objectmllm} serializes per-frame bounding box coordinates as text tokens and demonstrates significant improvements on video QA, particularly on motion-related questions, while Vamos~\cite{wang2024vamos} incorporates object state descriptions for versatile video understanding.
However, the token cost of text-coordinate approaches scales with both the number of objects and frames, creating the efficiency bottleneck discussed in \cref{sec:intro}.

An alternative strategy is to modify the visual input itself rather than encoding spatial information as text.
Shtedritski \etal~\cite{shtedritski2023redcircle} show that drawing a red circle on an image effectively directs a VLM's attention to a target region without any training.
Set-of-Mark~\cite{yang2023setofmark} overlays numbered markers on image segments for spatial referencing in GPT-4V, and ViP-LLaVA~\cite{cai2024vipllava} trains an MLLM to understand arbitrary visual prompts such as arrows, boxes, and circles.
FOCUS~\cite{zhong2025focus} leverages internal MLLM representations to guide visual cropping for fine-grained VQA; Zhang \etal~\cite{zhang2025mllmslook} studied how MLLMs process small visual details through attention patterns; and V$^*$~\cite{wu2024vstar} proposes a guided visual search mechanism for efficient querying in high-resolution images.
While these visual prompting methods establish that visual modifications can convey spatial information effectively, they focus on static images and single-region highlighting.
BoxTuning extends this paradigm to the video domain with multi-object colored overlays and trajectory trails that encode temporal motion patterns, a type of spatiotemporal information that has no natural text-coordinate analogue.

\section{Method}
\label{sec:method}

\begin{figure}[h]  \vspace{-10pt}
  \centering
  \includegraphics[width=0.8\linewidth]{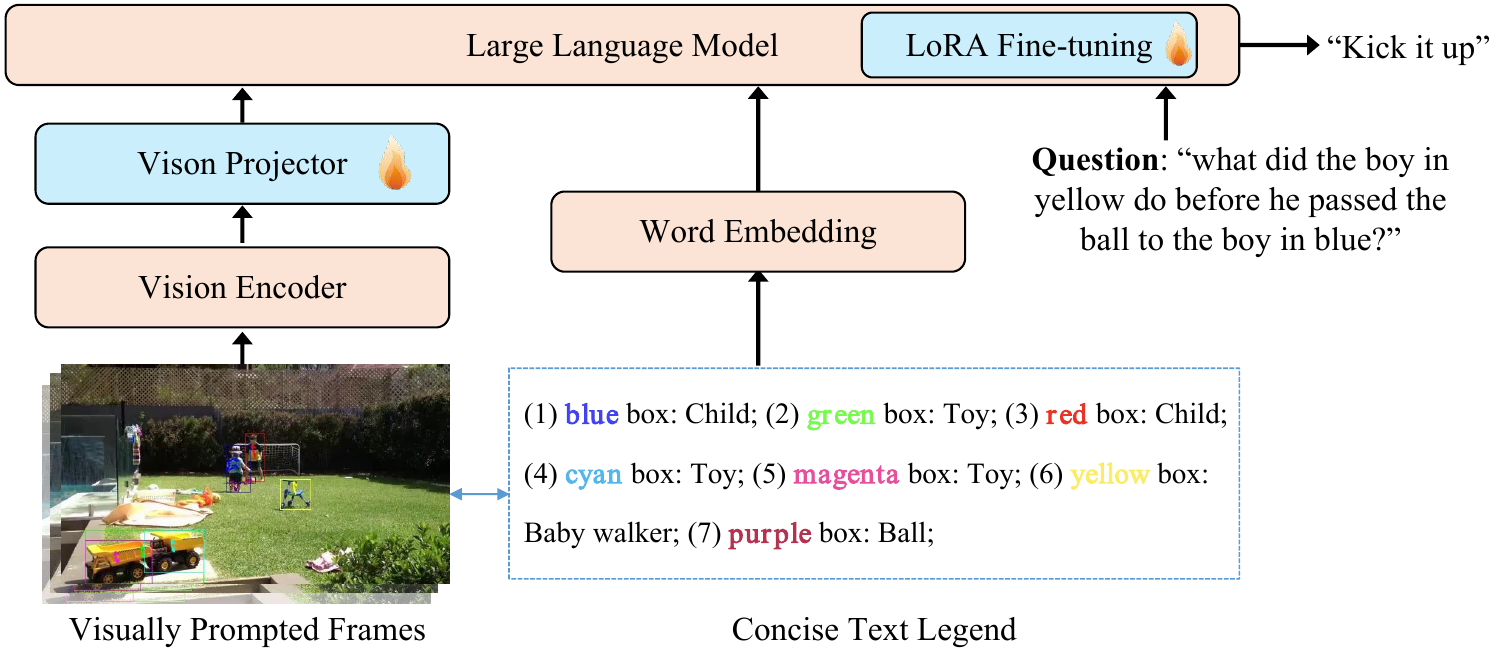}  \vspace{-5pt}
  \caption{Overview of the BoxTuning framework. Visually prompted frames, produced by rendering colored bounding boxes and trajectory trails from an off-the-shelf detector (YOLO-World~\cite{cheng2024yoloworld}) and tracker (SAM~2~\cite{ravi2025sam2}), are fed into the vision encoder together with a concise text legend mapping colors to object classes and the question. The vision encoder is kept frozen, while the vision projector (STC connector) is fully fine-tuned and the LLM is adapted via LoRA~\cite{hu2022lora}.}
  \label{fig:pipeline}  \vspace{-10pt}
\end{figure}

We aim to enhance video MLLMs with explicit object-level spatial-temporal information while avoiding the token overhead of text-coordinate representations.
To this end, we propose BoxTuning, a framework that renders object bounding boxes and motion trajectories as visual prompts directly on video frames.
\cref{fig:pipeline} illustrates the overall pipeline.
We first formulate the problem and contrast our approach with the text-coordinate paradigm (\cref{sec:formulation}), then describe the object detection and tracking pipeline (\cref{sec:detection}), the visual prompt construction (\cref{sec:overlay}), the concise text legend (\cref{sec:text}), and the fine-tuning procedure (\cref{sec:finetune}).
Implementation details are provided in \cref{sec:impl}.

\subsection{Problem Formulation}
\label{sec:formulation}

Consider a video QA task where a multimodal model must produce an answer $A$ given an input video $V = \{v_1, v_2, \ldots, v_T\}$ of \textit{$T$ uniformly sampled frames} and a text question $Q$.
We enhance the model with explicit object-level spatial information: an off-the-shelf detector and tracker (\cref{sec:detection}) provide \textit{$N$ tracked objects}, each with a class label $l_i$ and per-frame bounding boxes.
The key design choice is \textit{through which manner} the spatial-temporal information (from the detector and tracker) reaches the multimodal model.

ObjectMLLM~\cite{tang2025objectmllm} conveys all spatial-temporal information through text.
Each object's coordinates across frames are serialized into a token sequence:
\begin{quote}
\small
\texttt{$l_i$: frame\,1 [$x^1_1$\,$y^1_1$\,$x^1_2$\,$y^1_2$], frame\,2 [$x^2_1$\,$y^2_1$\,$x^2_2$\,$y^2_2$] \ldots frame\,T [$x^T_1$\,$y^T_1$\,$x^T_2$\,$y^T_2$]},
\end{quote}
where coordinates are quantized to $[0, 100]$.
The concatenation of all such strings is appended to the prompt alongside $Q$, incurring a text cost of $\mathcal{O}(N \times T \times C)$ tokens (where $C$ denotes the token requirement for the object's coordinates, $C{\approx}16$-$22$ per entry), while the video frames remain unmodified.

The proposed BoxTuning instead conveys spatial-temporal information directly through the visual modality.
We render colored bounding boxes and trajectory trails directly onto the video frames, producing augmented frames $\hat{V} = \{\hat{v}_1, \ldots, \hat{v}_T\}$.
Compared with ObjectMLLM~\cite{tang2025objectmllm} using $V$, the computational load of visual embedding does not increase.

The only text required is a concise legend $D$ that maps each color to its object class.
The model input thus becomes:
\begin{equation}
  \text{Input} = \bigl[\hat{V};\; D;\; Q\bigr],
  \label{eq:input}
\end{equation}
reducing the text cost from $\mathcal{O}(N \times T \times C)$ to $\mathcal{O}(N \times C')$, where $C'$ denotes the token requirement of $D$ for each object.
The following subsections describe the construction of $\hat{V}$ and $D$.

\subsection{Object Information Acquisition}
\label{sec:detection}

To obtain object bounding boxes across video frames, we employ a detect-then-track strategy using two complementary off-the-shelf models, following the same detection and tracking pipeline as ObjectMLLM~\cite{tang2025objectmllm} to ensure a fair comparison.

\noindent\textbf{Open-vocabulary detection.}
We use YOLO-World~\cite{cheng2024yoloworld}, an open-vocabulary object detector that accepts arbitrary category names as text prompts.
For each benchmark, we define a category vocabulary covering the object types relevant to that dataset.
Detection is performed on uniformly sampled keyframes, followed by non-maximum suppression to remove duplicate detections.

\noindent\textbf{Mask-based tracking.}
Initial detections are propagated across all frames using SAM~2~\cite{ravi2025sam2} (Segment Anything Model~2), which provides robust mask-based video object segmentation.
Given the bounding box prompts from the initial detections, SAM~2 generates per-frame segmentation masks that are robust to occlusions, scale changes, and appearance variations.
The per-frame bounding box $b_i^t{=}(x_1^t, y_1^t, x_2^t, y_2^t)$ for object $o_i$ in frame $v_t$ is derived as the tightest axis-aligned rectangle enclosing the predicted mask.

\noindent\textbf{Iterative detection for new objects.}
Objects may enter the scene after the first frame.
To handle this, we run YOLO-World on multiple uniformly spaced keyframes throughout the video.
For each subsequent keyframe, we compute the IoU between new detections and existing tracks; detections with high IoU against any existing track are discarded as duplicates, and the remaining ones initialize new tracks via SAM~2.
This iterative strategy ensures comprehensive object coverage across the entire video.

\subsection{Visual Prompt Construction}
\label{sec:overlay}

\begin{figure}[tb]
  \centering
  \begin{subfigure}{0.48\linewidth}
  	 \centering
    \includegraphics[width=0.85\linewidth]{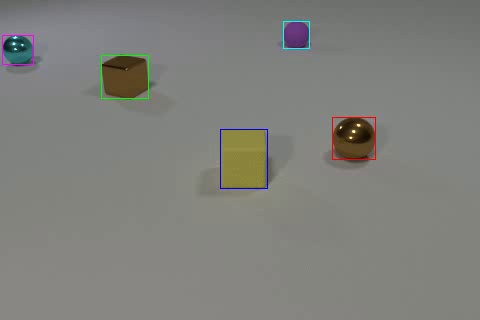}
    \caption{Box only}
    \label{fig:overlay-box}
  \end{subfigure}
  \hfill
  \begin{subfigure}{0.48\linewidth}
  	  \centering
    \includegraphics[width=0.85\linewidth]{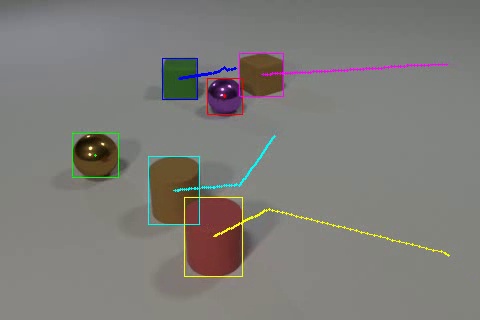}
    \caption{Box + trajectory trail}
    \label{fig:overlay-trail}
  \end{subfigure}  \vspace{-5pt}
  \caption{Visual prompt examples on CLEVRER video frames. (a)~Bounding boxes alone provide spatial localization of each object. (b)~Adding trajectory trails encodes motion direction and speed across intermediate frames, aiding temporal reasoning. The text legend (not shown) maps each color to its object class.}
  \label{fig:overlay} \vspace{-10pt}
\end{figure}

The core of BoxTuning is the construction of visual prompts that encode object spatial-temporal information directly on the video frames.
We compose two complementary elements: colored bounding boxes for spatial localization and trajectory trails for motion encoding, rendered with a consistent color scheme across all frames.

\noindent\textbf{Color assignment.}
Each tracked object $o_i$ is assigned a unique color $c_i$ from a predefined palette of $K$ maximally distinguishable colors (\eg, \textcolor{red}{red}, \textcolor{blue}{blue}, \textcolor{green}{green}, \textcolor{orange}{orange}, \textcolor{purple}{purple}, \textcolor{cyan}{cyan}, \ldots), as illustrated in \cref{fig:overlay}(a).
The same color is used consistently for a given object across all frames, enabling the model to associate the same object across time through color consistency.

\noindent\textbf{Bounding boxes for spatial localization.}
For each object $o_i$ visible in frame $v_t$, we draw a colored rectangle with color $c_i$ around its bounding box $b_i^t$.
The box provides precise per-frame spatial localization, allowing the model to directly `see' where each object is, without parsing coordinate numbers from text.

\noindent\textbf{Trajectory trails for motion encoding.}
For the video understanding task, temporal information is critical. Therefore, besides the object box for each frame, next we consider encoding the temporal information of the objects.
For video object perception, the motion trajectory is a commonly used representation of temporal information~\cite{luo2021multiple}.
In classical MLLM-based video understanding methods, only the sampled $T$ frames are used for visual embedding.
This way, to encode temporal motion trajectory, we draw a polyline connecting the bounding box centers of object $o_i$ over the most recent $L$ consecutive original frames
\begin{equation}
  \text{trail}_i^t = \bigl\{(\bar{x}_i^{t-L}, \bar{y}_i^{t-L}),\; \ldots,\; (\bar{x}_i^{t-1}, \bar{y}_i^{t-1})\bigr\},
  \label{eq:trail}
\end{equation}
where $(\bar{x}_i^t, \bar{y}_i^t)$ denotes the center of $b_i^t$, as in \cref{fig:overlay}(b).
The trail length $L$ is set equal to the sampling interval between two consecutive model input frames, \ie, $L = \lfloor T_{\text{video}} / T \rfloor$, where $T_{\text{video}}$ is the total number of original video frames and $T$ is the number of uniformly sampled frames.
This design ensures that the trail on each sampled frame covers exactly the temporal gap to the next sampled frame, so consecutive frames have adjacent trails that together reconstruct the complete motion trajectory without gaps.

Trajectory trails serve a purpose analogous to motion vectors in video codecs: they encode the direction and speed of object movement as a visual pattern rather than numerical values.
This is particularly valuable for causal and predictive reasoning, \eg, `Will the ball hit the cylinder?', where understanding motion direction is essential.

The augmented frame is obtained by compositing all overlays onto the original frame:
\begin{equation}
  \hat{v}_t = \text{Render}\bigl(v_t,\; \{(b_i^t, c_i, \text{trail}_i^t)\}_{i=1}^{N_t}\bigr),
  \label{eq:render}
\end{equation}
where $N_t$ is the number of visible objects in frame $v_t$. The trail is drawn in color $c_i$ from the box $b_i^t$, making the motion history of each object explicitly visible within a single frame.
The rendering is performed to preserve the underlying visual content.

\subsection{Concise Text Legend}
\label{sec:text}

While the visual prompts encode \textit{where} objects are and \textit{how} they move, the model still needs to know \textit{what} each colored object is.
We provide this information through a concise text legend:
\begin{quote}
\small
\texttt{(1) \textcolor{Blue}{blue} box: Child; (2) \textcolor{Green}{green} box: Toy; (3) \textcolor{red}{red} box: Child;~\ldots}
\end{quote}
where different colors can indicate the same category.
This legend requires only ${\sim}9$ tokens per object (index + color name + class label), compared to the ${\sim}18$ tokens \textit{per object per frame} required by the text-coordinate format.
\cref{tab:token_example} provides a concrete breakdown using NExT-QA as an example, with $N{\approx}8$ tracked objects. 
ObjectMLLM requires ${\sim}1{,}300$ text tokens for object spatial-temporal information, while BoxTuning's color legend requires only ${\sim}72$ tokens, a reduction of over 94\%.
The measured averages across all benchmarks are reported in \cref{tab:tokens} in the experiment section.

\begin{table}[htb!] \vspace{-10pt}
\scriptsize
  \caption{Token cost breakdown for a representative NExT-QA video ($N{\approx}8$ objects, $T{\approx}9$ temporally sampled frames per object).}
  \label{tab:token_example}
  \centering \vspace{-10pt}
  \begin{tabular}{lcc}
    \toprule
    Component & ObjectMLLM & BoxTuning (Ours) \\
    \midrule
    Per-entry cost   & ${\sim}18$ tok (frame\,+\,coords) &\quad ${\sim}9$ tok (index\,+\,label\,+\,color) \\
    Entries per video    & $N \times T \approx 8 \times 9 = 72$ & $N \approx 8$ \\
    \midrule
    \textbf{Total object text tok} & ${\sim} 18 \times 72 = \textbf{1{,}296}$ & ${\sim} 9 \times 8 =$ \textbf{72} \\
    \bottomrule
  \end{tabular}\vspace{-10pt}
\end{table}

This token reduction has two practical benefits.
1) \textit{Embedding space saving}: the token overhead is only marginally larger than the base MLLM without object input, leaving more capacity for video and question tokens.
2) \textit{Temporal downsampling needless}: since the text tokens in our prompting manner are not related to the number of frames, we can annotate every sampled frame at full temporal resolution, preserving fine-grained motion cues that text-coordinate methods~\cite{tang2025objectmllm} are forced to discard.

\subsection{Fine-Tuning Strategy}
\label{sec:finetune}

We adopt VideoLLaMA2-7B~\cite{cheng2024videollama2} as the base model, which consists of a CLIP ViT-L/14 vision encoder, a spatial-temporal convolution (STC) connector (vision projector) that captures cross-frame dependencies, and a Mistral-7B~\cite{jiang2023mistral} language model.
As shown in \cref{fig:pipeline}, we keep the vision encoder frozen and fine-tune the STC connector with full-parameter updates together with the language model via LoRA~\cite{hu2022lora}, using the same learning rate for both.
This design keeps the visual features intact while allowing the connector and language model to jointly learn to interpret the colored overlays and trajectory trails.

The training objective is the standard autoregressive next-token prediction loss on the answer tokens:
\begin{equation}
  \mathcal{L} = -\sum_{j=1}^{|A|} \log P_\theta\bigl(a_j \mid \hat{V}, D, Q, a_{<j}\bigr),
  \label{eq:loss}
\end{equation}
where $A = (a_1, \ldots, a_{|A|})$ is the ground-truth answer, $\theta$ includes the STC connector and LoRA adapter parameters.
For fair comparison, we use the same training data splits and evaluation protocols as ObjectMLLM~\cite{tang2025objectmllm} on all benchmarks.


\subsection{Implementation Details}
\label{sec:impl}

\noindent\textbf{Detection and tracking.}
We use YOLO-World-L~\cite{cheng2024yoloworld} for open-vocabulary detection with benchmark-specific category vocabularies (\eg, geometric shapes for CLEVRER; people, vehicles, and everyday objects for NExT-QA).
The confidence threshold is set to 0.3, and non-maximum suppression is applied at IoU~0.5.
Tracking uses SAM~2~\cite{ravi2025sam2} with default hyperparameters.
For iterative detection, new detections with IoU~$>$~0.5 against existing tracks are discarded as duplicates.

\noindent\textbf{Visual prompt settings.}
The color palette contains $K{=}32$ maximally distinguishable colors.
Bounding boxes are drawn with a line width of 3 pixels, and trajectory trails with 2 pixels.
Since we uniformly sample $T{=}16$ frames per video clip, the trail length $L = \lfloor T_{\text{video}} / T \rfloor$ varies across datasets depending on the original video clip length.

\noindent\textbf{Training.}
The STC connector is fully fine-tuned, and LoRA is applied to the query and value projection matrices of the language model with rank $r{=}128$, scaling factor $\alpha{=}256$, and dropout 0.05.
Both share the same learning rate.
We train with AdamW (learning rate $2{\times}10^{-4}$, weight decay 0.01, cosine schedule with 100 warmup steps) and batch size 16 on 4$\times$A100 80\,GB GPUs.

\section{Experiments}
\label{sec:exp}

\subsection{Evaluation Benchmarks}
\label{sec:benchmarks}

We include five video QA benchmarks spanning synthetic and real-world data:

\noindent\textbf{CLEVRER}~\cite{yi2020clevrer} contains synthetic videos of moving objects with questions about existence, motion direction, counting, attributes, and counterfactual inference.
Following ObjectMLLM~\cite{tang2025objectmllm}, we use the multi-choice subset (CLEVRER-MC) converted by MVBench~\cite{li2024mvbench}.

\noindent\textbf{Perception Test}~\cite{patraucean2023perception} is a diagnostic benchmark evaluating memory, abstraction, physics, and semantics across video and audio.

\noindent\textbf{STAR}~\cite{wu2021star} tests situated reasoning about interactions, sequences, predictions, and feasibility in real-world activity videos.

\noindent\textbf{NExT-QA}~\cite{xiao2021nextqa} targets causal (why/how), temporal (when/before/after), and descriptive (what) reasoning in videos. The emphasis on causal and temporal inference means that high-level action understanding often matters more than precise object localization.

\noindent\textbf{IntentQA}~\cite{li2023intentqa} evaluates understanding of human intent behind actions in videos. Like NExT-QA, it is reasoning-centric, requiring the model to infer \textit{why} and \textit{how} people act rather than simply locating objects.

\begin{table}[htb!] \vspace{-0pt}
	\caption{Average text tokens for object information per sample. ObjectMLLM~\cite{tang2025objectmllm} serializes bounding box coordinates as text. BoxTuning uses only a concise color legend.} \vspace{-10pt}
	\label{tab:tokens}
	\centering
	\scriptsize
	\begin{tabular}{lccc}
		\toprule
		Benchmark & ObjectMLLM & BoxTuning & Reduction \\
		\midrule
		CLEVRER         & 611   & 61   & 90.1\% \\
		Perception Test & 1{,}241  & 120  & 90.4\% \\
		STAR            & 761   & 96   & 87.4\% \\
		NExT-QA         & 1{,}184  & 77   & 93.5\% \\
		IntentQA        & 1{,}204  & 85   & 93.0\% \\
		\midrule
		Average         & 1{,}000  & 88   & 90.8\% \\
		\bottomrule
	\end{tabular}  \vspace{-5pt}
\end{table}

\subsection{Token Efficiency Analysis}
\label{sec:token}

A central claim of BoxTuning is the reduction in text token cost.
\cref{tab:tokens} quantifies this across all five benchmarks.
ObjectMLLM~\cite{tang2025objectmllm} requires 611-1,241 text tokens per sample for bounding box descriptions even after temporal downsampling, consuming the majority of the available text budget.
BoxTuning requires only 61-120 tokens regardless of the number of frames, achieving 87-93\% reduction.

\subsection{Main Results: BoxTuning vs.\ ObjectMLLM}
\label{sec:main_results}

\cref{tab:main} presents the main comparison to the base model without object information, and ObjectMLLM with text-coordinate bounding box encoding.
ObjectMLLM yields a large gain on CLEVRER (+9.7\%), where questions focus on object motion and collision, and modest gains on Perception Test (+0.6\%) and STAR (+0.7\%).
However, it \textit{hurts} performance on NExT-QA ($-$1.3\%) and IntentQA ($-$1.2\%), two reasoning-centric benchmarks where causal and intent inference matters more than precise object localization.
As ObjectMLLM~\cite{tang2025objectmllm} notes, these benchmarks focus on human actions and causal reasoning, which are difficult to capture with object bounding boxes alone. We further attribute the degradation to the token overhead: the verbose text coordinates crowd out the question and answer context due to token overflow, actively harming the model's reasoning capacity.

\begin{table}[h]
	\caption{Accuracy (\%) on five video QA benchmarks. All methods use VideoLLaMA2-7B with LoRA fine-tuning. Baseline numbers are from~\cite{tang2025objectmllm}. Best in \textbf{bold}. \textcolor{teal}{Green}/\textcolor{red}{red} indicate change relative to the base model.}
	\label{tab:main}
	\centering
	\scriptsize \vspace{-10pt}
	\begin{tabular}{lccccc}
		\toprule
		Method & CLEVRER & Perc.\ Test & STAR & NExT-QA & IntentQA \\
		\midrule
		VideoLLaMA2 (base)                & 67.9 & 66.0 & 66.5 & \textbf{79.8} & \textbf{76.7} \\
		+ ObjectMLLM (text coord.)        & 77.6 {\scriptsize\textcolor{teal}{+9.7}} & 66.6 {\scriptsize\textcolor{teal}{+0.6}} & 67.2 {\scriptsize\textcolor{teal}{+0.7}} & 78.5 {\scriptsize\textcolor{red}{-1.3}} & 75.5 {\scriptsize\textcolor{red}{-1.2}} \\
		+ BoxTuning (ours)                & \textbf{78.5} {\scriptsize\textcolor{teal}{+10.6}}  & \textbf{67.2} {\scriptsize\textcolor{teal}{+1.2}}  & \textbf{67.7} {\scriptsize\textcolor{teal}{+1.2}}  & 79.7 {\scriptsize\textcolor{red}{-0.1}}  & 76.5 {\scriptsize\textcolor{red}{-0.2}} \\
		\bottomrule
	\end{tabular} \vspace{-15pt}
\end{table}

BoxTuning outperforms ObjectMLLM on all five benchmarks while consuming only ${\sim}$10\% of the text tokens.
On the three spatially oriented benchmarks, BoxTuning achieves 78.5\% on CLEVRER (+0.9\% over ObjectMLLM), 67.2\% on Perception Test (+0.6\%), and 67.7\% on STAR (+0.5\%), demonstrating that visual prompts convey object spatial information more effectively than text coordinates even with drastically fewer tokens.

On the two reasoning-centric benchmarks, BoxTuning achieves 79.7\% on NExT-QA and 76.5\% on IntentQA, nearly matching the VideoLLaMA2 baseline (79.8\% and 76.7\%) with differences of only 0.1\% and 0.2\%, respectively.
In contrast, ObjectMLLM degrades the baseline by 1.3\% and 1.2\% on these two benchmarks.
This gap highlights a key advantage of our approach: BoxTuning's color legend occupies minimal text tokens, so the context window remains available for the question and reasoning content that these benchmarks demand.
The near-parity with the unmodified baseline further demonstrates the generalization of BoxTuning: injecting object information through visual prompts does not interfere with the model's reasoning capability, whereas the text-coordinate approach actively harms it by saturating the context budget.

\subsection{Comparison with Existing Video MLLMs}
\label{sec:comparison}

\cref{tab:sota} places BoxTuning in the broader landscape of video MLLMs.
BoxTuning achieves the highest accuracy on all five benchmarks, outperforming not only its direct baseline ObjectMLLM but also other competitive methods such as SeViLA and ViLA.
Notably, the zero-shot VideoLLaMA2 and LLaVA-NeXT-Video lag far behind on CLEVRER (45.6\% and 38.4\%), highlighting the difficulty of spatiotemporal object reasoning without explicit object information.
BoxTuning closes this gap effectively, reaching 78.5\% with the same VideoLLaMA2 backbone through visual prompting alone.

\begin{table}[h] \vspace{-15pt}
	\caption{Comparison with existing video MLLMs. Accuracy (\%) is reported.} \vspace{-10pt}
	\label{tab:sota}
	\centering
	\scriptsize
	\setlength{\tabcolsep}{3.5pt}
	\begin{tabular}{lcccccc}
		\toprule
		Method & Size & CLEVR. & Perc.T. & STAR & NExT & Intent \\
		\midrule
		LLaVA-NeXT-Vid.~\cite{zhang2024llavanextvideo}  & 7B & 38.4$^{*\dagger}$ & 49.3$^*$ & --   & --   & -- \\
		VideoLLaMA2~\cite{cheng2024videollama2}          & 7B & 45.6$^{*\dagger}$     & 51.4$^*$ & 57.1$^{*\dagger}$ & 74.1$^{*\dagger}$ & 73.8$^{*\dagger}$ \\
		SeViLA~\cite{yu2023sevila}                       & 3B & --                  & 62.0     & 64.9 & 73.8 & -- \\
		ViLA~\cite{wang2024vila}                         & 3B & --                  & --       & 67.1 & 75.6 & -- \\
		ObjectMLLM~\cite{tang2025objectmllm}             & 7B & 77.6                & 66.6     & 67.2 & 78.5 & 75.5 \\
		\midrule
		BoxTuning (ours)                                  & 7B & \textbf{78.5}     & \textbf{67.2}  & \textbf{67.7}  & \textbf{79.7}  & \textbf{76.5} \\
		\bottomrule
	\end{tabular} \\ \scriptsize{\leftline{\qquad \qquad \quad $^*$Zero-shot results. $^\dagger$Reproduced by~\cite{tang2025objectmllm}.}} \vspace{-20pt}
\end{table}

\subsection{Ablation Studies}
\label{sec:ablation}


\subsubsection{Component-wise Ablation.}
\cref{tab:ablation_trail} isolates the contribution of each BoxTuning component by progressively adding the text legend and trajectory trails to the bounding box baseline.
Bounding boxes alone already achieve accuracy comparable to ObjectMLLM (\eg, 77.0\% \vs 77.6\% on CLEVRER), confirming that visual prompts can convey spatial information nearly as well as text coordinates.
Adding the text legend brings a small but consistent improvement across all benchmarks (+0.2--0.4\%), as the explicit color-to-class mapping helps the model associate visual cues with object identities.
The largest gain comes from trajectory trails: on motion-heavy benchmarks, accuracy jumps by +1.2\% on CLEVRER, +0.7\% on Perception Test, and +0.5\% on STAR, showing that the trails effectively recover inter-frame motion dynamics.
On the reasoning-centric NExT-QA and IntentQA, the improvement is smaller (+0.3\%) but still positive, indicating that trajectory trails do not interfere with causal reasoning while providing modest additional context.

\begin{table}[htb!] \vspace{-15pt}
	\caption{Component-wise ablation. Accuracy (\%) is reported.} \vspace{-10pt}
	\label{tab:ablation_trail}
	\centering
	  \scriptsize
	\begin{tabular}{lccccc}
		\toprule
		Variant & CLEVRER & Perc.\ Test & STAR & NExT-QA & IntentQA \\
		\midrule
		Box only                & 77.0  & 66.3  & 66.8  & 79.3  & 76.1 \\
		Box + Legend            & 77.3  & 66.5  & 67.2  & 79.4  & 76.2 \\
		Box + Legend + Trail    & 78.5  & 67.2  & 67.7  & 79.7  & 76.5 \\
		\bottomrule
	\end{tabular} \vspace{-20pt}
\end{table}

\subsubsection{Effect of Trajectory Length.}
We vary the trail length using fixed $L \in \{10, 20, 30, 40, 60\}$ and compare against our adaptive length ($L = \lfloor T_{\text{video}} / T \rfloor$) on two motion-sensitive benchmarks: CLEVRER and Perception Test.
As shown in \cref{tab:ablation_length}, short trails ($L{=}10$) lead to a notable accuracy drop as they fail to capture sufficient motion context, while longer trails ($L{\geq}30$) yield similar performance regardless of the exact value.
The adaptive strategy ($L = \lfloor T_{\text{video}} / T \rfloor$) outperforms all fixed lengths, as it aligns the trail length with the actual sampling interval, thereby encoding the complete trajectory between adjacent sampled frames without redundancy or truncation.
This demonstrates that the adaptive length obtains the optimal trade-off between motion context gain and visual clutter distraction.

\begin{table}[htb!] \vspace{-10pt}
	\caption{Ablation study on trajectory trail length $L$. Accuracy (\%) is reported.} \vspace{-10pt}
	\label{tab:ablation_length}
	\centering
	\scriptsize
	\begin{tabular}{lcccccc}
		\toprule
		Trail length $L$ & 10 & 20 & 30 & 40 & 60 & Adaptive \\
		\midrule
		CLEVRER        & 77.6  & 78.0  & 78.2  & 78.1  & 78.0  & \textbf{78.5} \\
		Perc.\ Test    & 66.5  & 66.9  & 67.0  & 67.0  & 66.9  & \textbf{67.2} \\
		\bottomrule
	\end{tabular} \vspace{-25pt}
\end{table}

\subsubsection{Effect of Line Width.}
We study the sensitivity to the rendering line width for bounding boxes and trajectory trails, separately, on CLEVRER.
As shown in \cref{tab:ablation_width}, a box width of 3 and a trail width of 2 achieve the best accuracy.
Thinner lines ($w{=}1$) are less visible at the model's input resolution, weakening the spatial signal, while wider lines ($w{\geq}5$) occlude the underlying frame content and degrade performance noticeably.

\begin{table}[htb!]  \vspace{-10pt}
  \caption{Ablation study on rendering line width on CLEVRER using accuracy (\%).} \vspace{-10pt}
  \label{tab:ablation_width}
  \centering
  \scriptsize
  \begin{tabular}{ccccccc}
    \toprule
     Width & 1 & 2 & 3 & 5 & 8 \\
    \midrule
    Box    & 77.8 & 78.2 & \textbf{78.5} & 78.0 & 77.1 \\
    Trail  & 78.0 & \textbf{78.5} & 78.3 & 77.4 & 76.8 \\
    \bottomrule
  \end{tabular} \vspace{-20pt}
\end{table}

\section{Conclusion}
\label{sec:conclusion}

We have designed BoxTuning, a visual prompting approach that encodes object bounding boxes and trajectory trails directly on video frames, replacing the text-coordinate paradigm with a visually grounded alternative.
By shifting object information from the text modality to the visual modality, BoxTuning reduces text token cost by 87\%-93\% and eliminates the need for temporal downsampling, preserving fine-grained motion cues that text-coordinate methods are forced to discard.
Experiments on five video QA benchmarks show that BoxTuning exceeds text-coordinate baselines on spatially oriented tasks while nearly eliminating the accuracy degradation on reasoning-centric benchmarks caused by text token overflow.
Ablation studies confirm the contribution of trajectory trails for motion-sensitive tasks, validate an adaptive trail length strategy that encodes complete inter-frame trajectories, and show that the rendering hyperparameters are robust within a reasonable range.
We hope these findings encourage further exploration of visual prompting as a general strategy for conveying structured information to multimodal models.


\bibliographystyle{splncs04}
\bibliography{main}
\end{document}